\documentclass[letterpaper, 10pt, conference]{ieeeconf}  

\IEEEoverridecommandlockouts                              

\usepackage{amsmath} 
\usepackage{amssymb}  

\usepackage{algorithmic}
\usepackage{algorithm}
\usepackage{array}
\usepackage[caption=false,font=normalsize,labelfont=sf,textfont=sf]{subfig}
\usepackage{textcomp}
\usepackage{stfloats}
\usepackage{url}
\usepackage{verbatim}
\usepackage{graphicx}
\usepackage{cite}
\usepackage{color}
\graphicspath{{./images/}}
\usepackage{mathtools,stackengine}
\stackMath

\usepackage{tabularx,booktabs}
\newcolumntype{C}{>{\centering\arraybackslash}X} 
\usepackage{diagbox}

\usepackage{subfloat}
\newbox{\bigpicturebox}
\usepackage[export]{adjustbox} 

\usepackage{bm}

\usepackage{soul}

\title{\LARGE \bf
Surgical Task Automation Using Actor-Critic Frameworks and Self-Supervised Imitation Learning*
}

\author{Jingshuai Liu$^{1}$, Alain Andres$^{2}$, Yonghang Jiang$^{3}$, Xichun Luo$^{3}$, Wenmiao Shu$^{4}$, and Sotirios A. Tsaftaris$^{1}$
\thanks{*This work has been submitted to the IEEE for possible publication. Copyright may be transferred without notice, after which this version may no longer be accessible.}
\thanks{*This work was supported by Engineering and Physical Sciences Research Council (EPSRC), Project Real-time Digital Twin Assisted Surgery, EP/X033686/1.}
\thanks{$^{1}$Jingshuai Liu and Sotirios Tsaftaris are with the Institute for Imaging, Data and Communications, School of Engineering, University of Edinburgh, Edinburgh EH9 3JE, U.K. (e-mail: jliu11@ed.ac.uk; s.tsaftaris@ed.ac.uk). (Corresponding author: Sotirios Tsaftaris)}
\thanks{$^{2}$ Alain Andres is with TECNALIA, Basque Research and Technology Alliance (BRTA), Mikeletegi Pasealekua 2, Donostia-San Sebastian, 20009, Spain. (e-mail: alain.andres@tecnalia.com).}
\thanks{$^{3}$ Yonghang Jiang and Xichun Luo are with Centre for Precision Manufacturing, Department of Design, Manufacturing and Engineering Management, University of Strathclyde, Glasgow G1 1XJ, U.K. (e-mail: yonghang.jiang@strath.ac.uk; xichun.luo@strath.ac.uk).}
\thanks{$^{4}$ Wenmiao Shu is with Department of Biomedical Engineering, University of Strathclyde, Glasgow G4 0NW, U.K. (e-mail: will.shu@strath.ac.uk).}
}

\begin{document}

\maketitle
\thispagestyle{empty}
\pagestyle{empty}

\begin{abstract}
    Surgical robot task automation has recently attracted great attention due to its potential to benefit both surgeons and patients. Reinforcement learning (RL) based approaches have demonstrated promising ability to provide solutions to automated surgical manipulations on various tasks. To address the exploration challenge, expert demonstrations can be utilized to enhance the learning efficiency via imitation learning (IL) approaches. However, the successes of such methods normally rely on both states and action labels. Unfortunately action labels can be hard to capture or their manual annotation is prohibitively expensive owing to the requirement for expert knowledge. It therefore remains an appealing and open problem to leverage expert demonstrations composed of pure states in RL. In this work, we present an actor-critic RL framework, termed AC-SSIL, to overcome this challenge of learning with state-only demonstrations collected by following an unknown expert policy. It adopts a self-supervised IL method, dubbed SSIL, to effectively incorporate demonstrated states into RL paradigms by retrieving from demonstrates the nearest neighbours of the query state and utilizing the bootstrapping of actor networks. We showcase through experiments on an open-source surgical simulation platform that our method delivers remarkable improvements over the RL baseline and exhibits comparable performance against action based IL methods, which implies the efficacy and potential of our method for expert demonstration-guided learning scenarios.
\end{abstract}


\section{Introduction}

Reinforcement learning (RL), which is a specialized branch within the broad field of artificial intelligence (AI) and provides a type of machine learning techniques for automated decision-making~\cite{RL_book,deep_RL}, has witnessed rapid advancements and impactful innovations in various domains, with medical surgical assistance being one prominent area of application~\cite{RL_surgery,RL_surgical_decision_making,survey_RL_medical}. In recent years, deep learning models have achieved great success in diverse domains via the spectacular learning capacity and rich representation of deep neural network architectures~\cite{alexnet,vgg,vallina_gan,vit,swin_transformer}. Reinforcement learning, empowered by advances in deep learning domains, is transforming medical surgical assistance and automated decision-making~\cite{DRL_robot_train,RL_robot_sugery,sim2real_autonomous,sim2real_CycleGAN} and the evolution of these technologies continuously enhance the accuracy, efficiency, and reliability of medical treatments and interventions~\cite{loop_healthcare,general_RL_Med,GAIL_RL_robot,DRL_robot_train}, in the pursuit of improved patient outcomes.

\begin{figure}[t]
    \begin{minipage}[b]{1.0\linewidth}
        \centering
        \centerline{\includegraphics[width=0.76\linewidth]{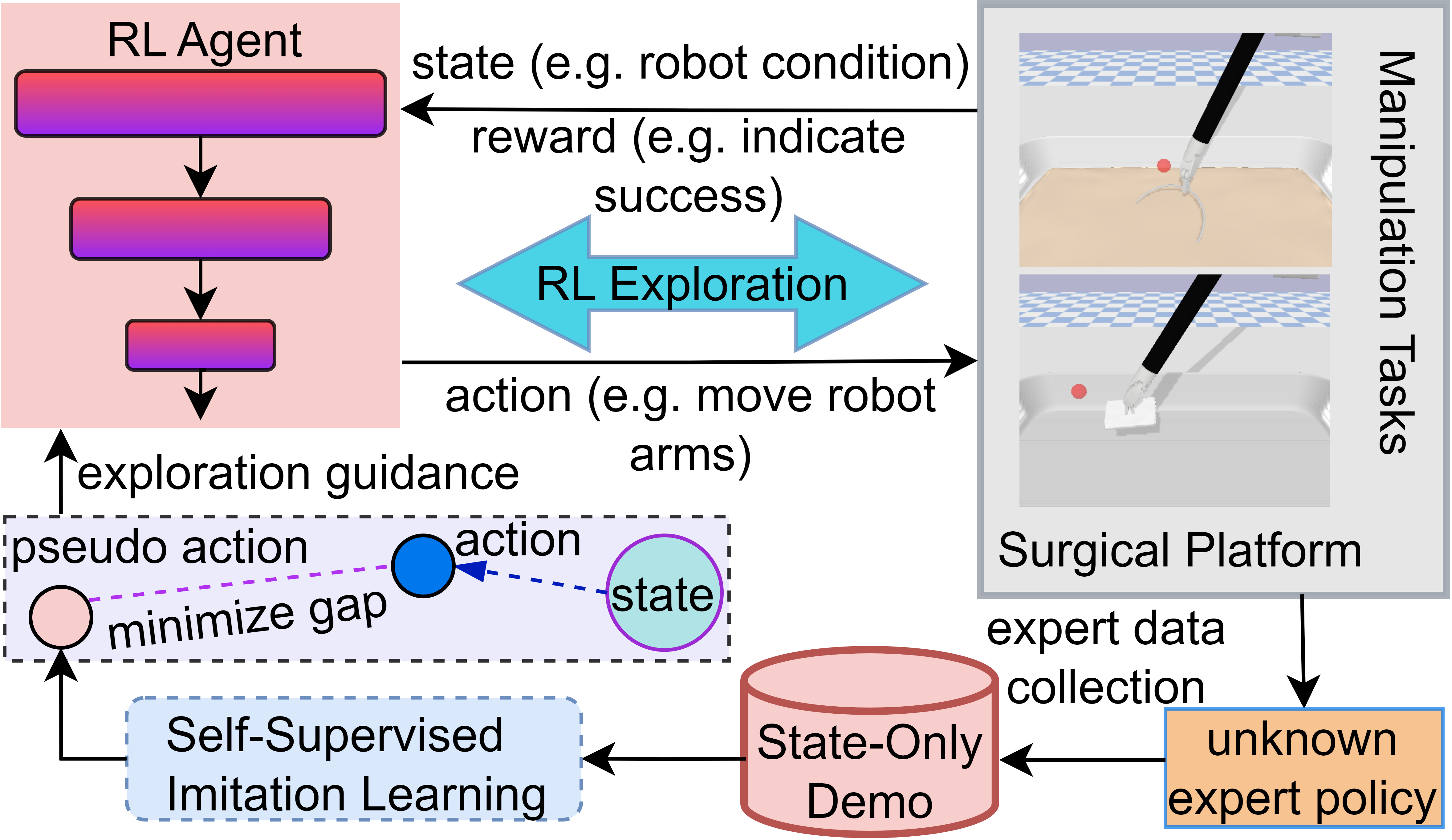}}
    \end{minipage}
    \caption{Our method learns to perform automated surgical tasks using reinforcement learning (RL) algorithms where an agent makes actions based on states in a simulated platform, receives feedback, and improves over time. We propose a novel self-supervised imitation learning approach to leverage state-only demonstrations, i.e. pure states without action information, collected by an unknown expert policy, in order to enhance RL exploration.}
    \label{fig:ACSIL_overall}
\end{figure}

The learning of RL models traditionally depends on large amounts of data collected via online interactions or extensive exploration to optimize the policy for decision-making. The quality and efficiency of model training is associated with the exploration capacity which can be improved by adopting the recipe of incorporating expert demonstrations into the learning process~\cite{DDPGBC,CoL,minimalist_offline,revisit_minimalist_offline}. Despite success achieved by imitation learning methods with expert demonstrations~\cite{survey_IL,survey_demo_robot}, it can be impossible to obtain actual actions of the expert in many scenarios. For instance, sensors used to record actions can be affected by noise and introduce inaccuracies that are detrimental to action label reliability. Moreover, demonstrated actions can be executed continuously and thus hard to be converted into actions suitable for robot learning. Manually annotating actions, especially in a detailed and precise manner, is prohibitively expensive and time-consuming, which particularly holds for complex robotic tasks that require expert knowledge.

It is thus intriguing and promising to explore avenues for utilizing demonstrations which are collected with an unknown expert policy and composed of pure states~\cite{IL_Obs_survel,RCE,LobsDICE}. We therefore propose a novel approach to incorporating state-only imitation learning into reinforcement learning paradigms. We then can drive the agent policy towards the demonstrated policy to improve the learning efficiency while maintaining the exploration efficacy to retrieve optimal solutions. As we detail in the related work section in Sect.~\ref{sect:related_works}, our method is different to other works that are trained in an adversarial framework or only perform guidance on action value approximation~\cite{GAIL,GAIfO,AMP,seabo}. Our method leverages state-only demonstrations to guide the agent exploration in a self-supervised manner, as demonstrated in Fig.~\ref{fig:ACSIL_overall}. Our method shows performance improvements and circumvents known issues with adversarial methods such as instability and limited generalization leading to performance degradation~\cite{GAIL,interpretable_GAIL,survey_IRL}.

In this paper, we are particularly interested in addressing the challenge of learning with expert demonstrations only consisting of states for surgical task automation. To tackle this challenge, we present an actor-critic RL framework, termed AC-SSIL, which adopts a novel self-supervised imitation learning method, dubbed SDIL, to guide the learning process by retrieving from demonstrations the nearest neighbours of the states visited by the agent and using the target actor network to produce pseudo action labels for exploration guidance. Different from self-imitation learning proposed in~\cite{SIL} to reproduce the past good experiences, our method utilizes expert demonstrations that \textit{only} have state observations to aid agent learning. The contributions of this work are summarized as follows:
\begin{itemize}
    \item An actor-critic RL framework, termed AC-SSIL, to learn policies for automated surgical tasks by incorporating expert demonstrations into RL paradigms to enhance model exploration;
    \item A self-supervised imitation learning method, dubbed SSIL, to leverage demonstration data consisting of pure states to improve the agent training while mitigating the necessity of action annotations;
    \item Our experiments, demonstrate that the proposed AC-SSIL yields significant improvements compared to the RL baseline, and outperforms or is on par with other existing approaches which rely on action labels. Our ablation studies assess the impact of algorithmic designs and show the efficacy of our method in improving model learning and its insensitivity to specific parameters.
\end{itemize}

\section{Related Works}
\label{sect:related_works}


In this section, we begin with a review of RL methods for surgical task automation and assistance, and then introduce methods for imitating expert behaviour which fall under one of two categories: learning with expert data including demonstrated actions and learning from pure observations.

\subsection{Reinforcement Learning for Surgical Assistance}

The evolution of reinforcement learning, a form of machine learning, has facilitated advancements in automated decision-making and assistance in medical surgeries and provided a powerful paradigm where an agent is trained via interactions with an environment to make decisions that can maximize the cumulative rewards, e.g. the success rate and patient recovery~\cite{survey_RL_medical,general_RL_Med,human_loop_surgical_sim}. Reinforcement learning approaches enable agents to acquire necessary skills to perform surgical tasks from data collected via online interactions or from previous trials, and have demonstrated improvements in model applicability, flexibility, and generalization capacity in automating surgical tasks~\cite{RL_robot_inverse,multi_agent_robot_surgery,IRL_steer_needle,organ_context_RL}. An RL framework for learning surgical manipulation skills was proposed in~\cite{RL_needle_pick}, which deploys an agent in am implicit curriculum learning scheme and leverages the knowledge of the prior critic via $Q$-value function transfer. To deal with long-horizon tasks in surgeries, an approach was introduced in~\cite{ViSkill} to divide a task into several sub-tasks and respectively train sub-task policies which are smoothly connected via a value-informed skill chaining method. Despite the promising progress achieved, executing RL methods in surgical tasks can still be challenging due to exploration problems, e.g. sample efficiency and the balance between exploration and exploitation, which potentially impact the performance of RL algorithms. We therefore develop a framework to guide and enhance model learning by leveraging a few expert demonstrations.


\subsection{Imitation Learning with Expert Demonstrations}

Expert demonstrations have proven useful in improving the exploration efficiency of RL models by emulating the expert behaviour via imitation learning (IL) approaches~\cite{survey_IL,survey_demo_robot} which show to help the model learn skills to accomplish complicated tasks. Behaviour cloning (BC)~\cite{DDPGBC,SQIL} emerged as a method for imitating expert behaviours from demonstrations by minimizing the distance between the actions proposed by the agent and the demonstrated actions, which can be implemented with offline datasets or incorporated into RL algorithms to offer a regularization for model learning. The work in~\cite{driving_online_IL} learned a deep neural network control policy for high-speed driving by applying IL to mimic an expert policy. The works in~\cite{minimalist_offline,revisit_minimalist_offline} explored the designs of adding BC term to RL and showed remarkable improvements for offline RL, owing to the behaviour-regularized actor-critic algorithms. In~\cite{CoL}, it was evidently demonstrated that the utilization of a few demonstration trajectories is conducive to the performance of RL models via the combination of the BC loss and the RL objective to facilitate the learning procedure. An advantage weighted actor-critic framework, termed AWAC~\cite{awac}, was proposed to transfer knowledge from previously collected experiences to prevent inefficient exploration and offer a less conservative training paradigm by re-weighting the objective via the estimated action values. The method introduced in~\cite{dex} guides model learning by incorporating demonstrations into agent exploration to improve training efficiency.

Although useful in practice, those protocols normally require actions in tandem with the demonstration data. Such actions can be expensive and implausible to collect and annotate. Hence such protocols cannot be applied in state-only regimes where only the environment observations are recorded in expert demonstrations and no action annotations are available. To extend demonstration guidance to state-only scenarios, we propose to guide RL exploration with pure states through a self-supervised imitation learning approach.

\subsection{Learning from Observations}

To apply imitation learning in state-only scenarios where expert demonstrations only contain states, state cloning (SC)~\cite{hybrid_IL} was proposed to enforce the agent to produce state transitions that are similar to those picked up from demonstrated trajectories~\cite{hybrid_IL}. A similar method performs behavior cloning using the actions inferred by an inverse dynamics model~\cite{BC_obs}. Although they offer an alternative to BC approaches, such methods involve training a separate model to capture the environment dynamics, which can cause training instability and lead to performance degradation when the prediction accuracy is limited, e.g. the environment changes are too complex to learn or the amount of samples is insufficient to develop a high-precision model. Additionally, they may require establishing a series of task-specific dynamics models when considering diverse tasks that have distinct state and action spaces and demand various capacities to model the environment dynamics. 

Alternative approaches that can leverage state-only demonstrations were introduced in~\cite{AMP,GAIfO}, which extend a generative adversarial imitation learning (GAIL) framework~\cite{GAIL} to optimize the policy agent towards producing state transitions that are indistinguishable from demonstration data. Inverse reinforcement learning (IRL) methods~\cite{IRL,entropy_IRL} were proposed to learn and improve upon the demonstrated behaviour by inferring a reward function from demonstrations which allows an agent to make decisions according to the inferred rewards. Although providing methods for imitation learning without explicit action labels, they can suffer from training instability and difficulties in convergence, which makes it challenging to effectively develop agents~\cite{GAIL,interpretable_GAIL}. The model performance can be significantly affected by the quality and quantity of demonstrations, particularly in complex environments with a large state space, and the learned policy might not generalize well to unseen situations when a large amount of examples that are adequately representative of the task space are unavailable~\cite{GAIL,survey_IRL,interpretable_GAIL}. Advances in reward engineering for RL were made to leverage demonstrated state trajectories to refine the reward function used to train the agent, by measuring the difference from demonstrations in the state space to regularize the action value approximation~\cite{OT_offlineRL,seabo}. Despite efficient implementations and efficacy in reshaping the reward function, those avenues aim to enhance the estimation of action values and provide no guidance on updating policy agents and penalizing actions deviating from the demonstrated behaviour. These factors impede the implementation of imitation learning techniques and motivate us to devise an effective and efficient method for learning from state-only demonstrations.

We develop a self-supervised method for utilizing demonstrated states to guide the exploration of the agent, while avoiding the issues of methods based on adversarial training or the refinement of action value approximation. Our method retrieves from demonstrations the nearest neighbours of the query state and bootstraps the learned actor network to provide demonstration guidance which shows to improve the agent performance. The recipe of learning from observations enables utilizing broad expert demonstration resources that only provide state information. 

\section{Methodology}



We develop a framework, as summarized in Fig.~\ref{fig:ACSIL_overall}, to achieve surgical task automation, which is established based on reinforcement learning (RL) algorithms where an agent learns manipulation skills by interacting with a simulated environment, i.e. take actions, get feedback, and improve over time. The learning process is enhanced with expert demonstrations where the exact actions taken by an unknown expert policy are unavailable and only states are collected. Our framework uses a self-supervised method to guide agent training without needing detailed action information in demonstrations which is typically necessary in traditional imitation learning (IL) approaches.

In the following, we formulate the problem of reinforcement learning with expert demonstrations consisting of pure states in Sect.~\ref{sect:learn_from_demo} and introduce the preliminaries of actor-critic RL in Sect.~\ref{sect:RL_preliminary}. The proposed self-supervised imitation learning method, termed SSIL, is described in Sect~\ref{sect:SIL} which is followed by the introduction of the actor-critic SSIL training framework, dubbed AC-SSIL, in Sect~\ref{sect:AC_SIL}.

\subsection{Problem Formulation: Learning with State-Only Demonstrations}
\label{sect:learn_from_demo}

\begin{figure*}[thpb]
    \begin{minipage}[b]{1.0\linewidth}
        \vspace{5pt}
        \centering
        \centerline{\includegraphics[width=0.65\linewidth]{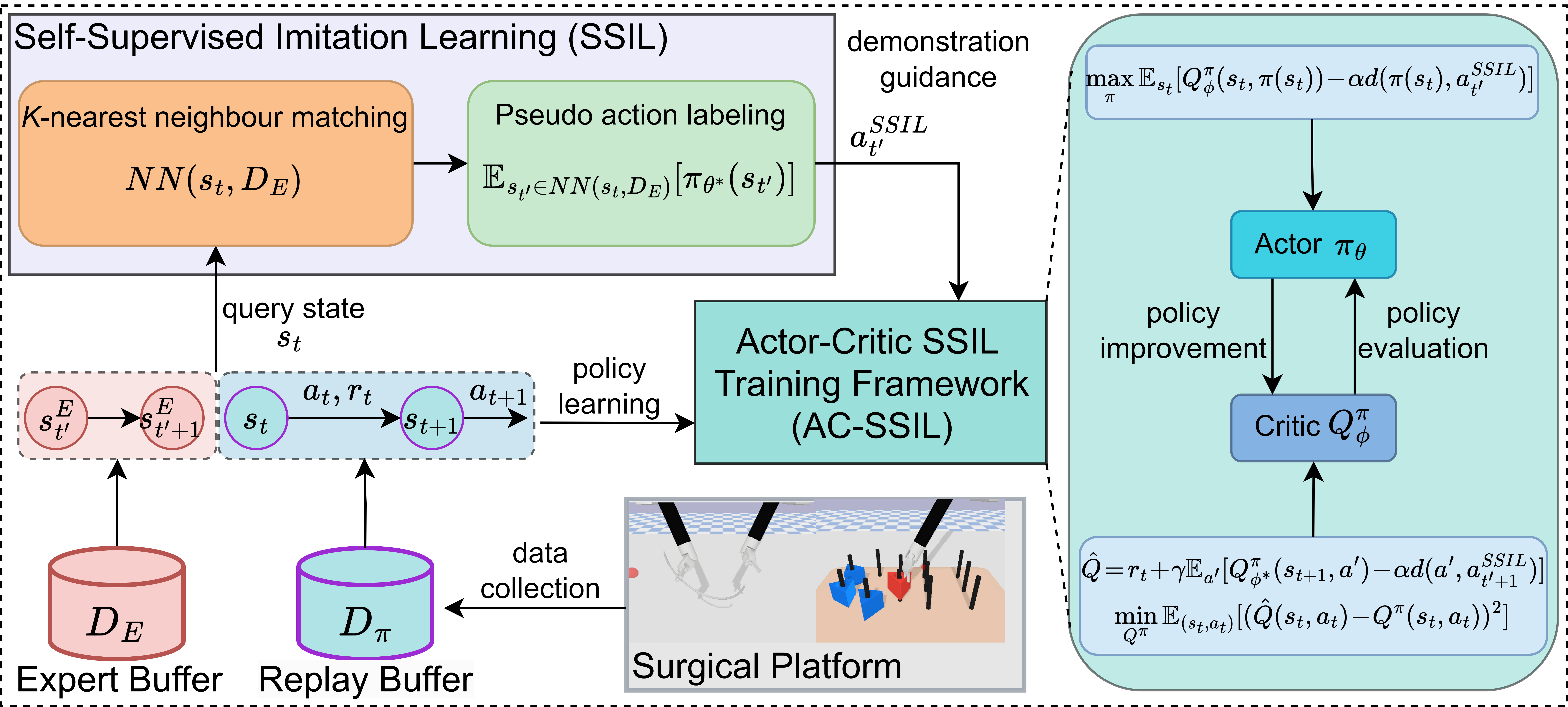}}
    \end{minipage}
    \caption{Illustration of actor-critic SSIL training framework (AC-SSIL). The replay buffer $D_\pi$ and the expert buffer $D_E$ are collected from the surgical platform and used to optimize the policy agent. Since no actions are available in $D_E$, the devised self-supervised imitation learning (SSIL) is adopted to provide guidance on model training in a reinforcement learning paradigm.}
    \label{fig:ACSIL_demo}
\end{figure*}

We address the policy learning in an online interactive environment which is formulated by a Markov decision process. The agent takes an action $a_t$ at the $t$-th time step based on the currently observed state $s_t$ and its policy $\pi$. The environment responds to the executed action by returning a reward, termed $r_t$, and transiting to the successive state $s_{t+1}$ which elicits action $a_{t+1}$ for the next time step. The action-making process and state transitions are stored as experimental experience into a replay buffer~\cite{HER}, dubbed $D_\pi$, in the form of tuples $\{(s_t,a_t,r_t,s_{t+1},a_{t+1})\}_t$. Meanwhile, experiences generated by an unknown expert policy are maintained in a expert demonstration buffer $D_E=\{(s_{t'}^E,s_{t'+1}^E)\}_{t'}$ with state-only recordings where action annotations and reward labels are inaccessible, as demonstrated in Fig.~\ref{fig:ACSIL_demo}. We utilize the replay buffer for policy optimization via reinforcement learning (RL) algorithms and propose to guide the learning process using state-only demonstrations. Since the expert buffer has no action labels, conventional RL and imitation learning approaches are inapplicable, which motivates us to develop the self-supervised imitation learning method for harnessing the demonstration knowledge to steer and enhance agent exploration in state-only scenarios.

\subsection{Deep Reinforcement Learning Fundamentals}
\label{sect:RL_preliminary} 

Within the spectrum of AI technologies, reinforcement learning algorithms are developed to make a decision on action selection when taking as input an observable or partially observable state. They are optimized to take a series of actions which are intricately coordinated to reach a desired state or maximize the cumulative reward. RL agents are trained on a dataset of tuples $\{(s_t,a_t,r_t,s_{t+1},a_{t+1})\}_t$ which are recorded online or collected from the previous trials, where $s$ denotes the observable state, $a$ refers to the action, $r$ is the reward, and $t$ denotes the time step. In our experiments, the state and action spaces are continuous and respectively composed of the object and robot states and Cartesian-space control. In a deep RL framework, an actor network is responsible for proposing actions by learning a policy that maps states to actions, and a critic network evaluates the quality of actions selected by the actor. Their combination can enhance the learning efficiency by leveraging both policy gradients and value function estimates~\cite{DQN}.

The goal of training the actor network is to maximize the cumulative reward defined as follows:
\begin{equation}
    \max\limits_{\theta} \mathbb{E}_{\pi_\theta}[\sum\limits_{t=0}^{\infty}\gamma^t r_t],
    \label{eq:cumulative_reward}
\end{equation}
where $\pi_\theta$ is the actor network and $\theta$ denotes its parameters, $\gamma$ is the discount factor, e.g. $\gamma=0.99$ as adopted in our experiments, and $\mathbb{E}$ denotes the expectation.

To provide a more efficient sampling and learning paradigm, $Q$-learning function approximation~\cite{q_learning} is adopted to estimate the accumulated reward via bootstrapping approaches where the current action value estimate is updated based on other estimates via temporal difference (TD). Neural network models are normally adopted for value estimation due to their excellent learning and generalization capacities, which forms deep $Q$-network algorithms~\cite{DQN}. This practice allows RL models to learn from partial sequences of data rather than complete episodes and leads to faster convergence in many scenarios~\cite{RL_book,RL_survey_robot,DQN}. The action value is estimated via a critic network $Q_\phi^\pi$ which is updated by:
\begin{equation}
    Q_\phi^\pi = \arg\min\limits_{Q^\pi} \mathbb{E}_{(s_t,a_t)}[(\hat{Q}(s_t, a_t) - Q^\pi(s_t, a_t))^2],
    \label{eq:ac_critic_loss}
\end{equation}
where $\phi$ denotes the parameters of the critic, tuples $(s_t,a_t,r_t,s_{t+1})$ are generated by taking a series of actions following the policy of actor $\pi_\theta$ and stored in a buffer, and $\hat{Q}$ is the target $Q$-value and computed via bootstrapping:
\begin{equation}
    \hat{Q}(s_t, a_t) \coloneqq r_t + \gamma \mathbb{E}_{a'\sim \pi_{\theta^\ast}}[Q_{\phi^\ast}^\pi(s_{t+1}, a')],
    \label{eq:ac_target_critic}
\end{equation}
where $\theta^\ast$ and $\phi^\ast$ are the parameters of the target actor and critic networks which are utilized to stabilize the learning process and alleviate the critical issue of overestimation, due to function approximation error which occurs when the estimated $Q$-values for taking a specific action are higher than their true values and can result in sub-optimal policies and slower learning~\cite{TD3,TATD3}. The target network parameters are updated via the exponential moving average:
\begin{align}
    \theta^\ast &\leftarrow \tau\theta + (1-\tau)\theta^\ast \notag\\
    \phi^\ast &\leftarrow \tau\phi + (1-\tau)\phi^\ast,
    \label{eq:target_networks}
\end{align}
where $\tau$ is the target update rate and normally set to be 0.005.

The objective of the actor is to optimize the policy of action-making to maximize the values estimated by the critic:
\begin{equation}
    \pi_\theta = \arg\max\limits_{\pi} \mathbb{E}_{s_t}[Q_\phi^\pi(s_t, \pi(s_t))].
    \label{eq:ac_actor_loss}
\end{equation}

\subsection{Self-Supervised Imitation Learning (SSIL)}
\label{sect:SIL}
\begin{figure}[thpb]
    \begin{minipage}[b]{1.0\linewidth}
        \centering
        \centerline{\includegraphics[width=1.0\linewidth]{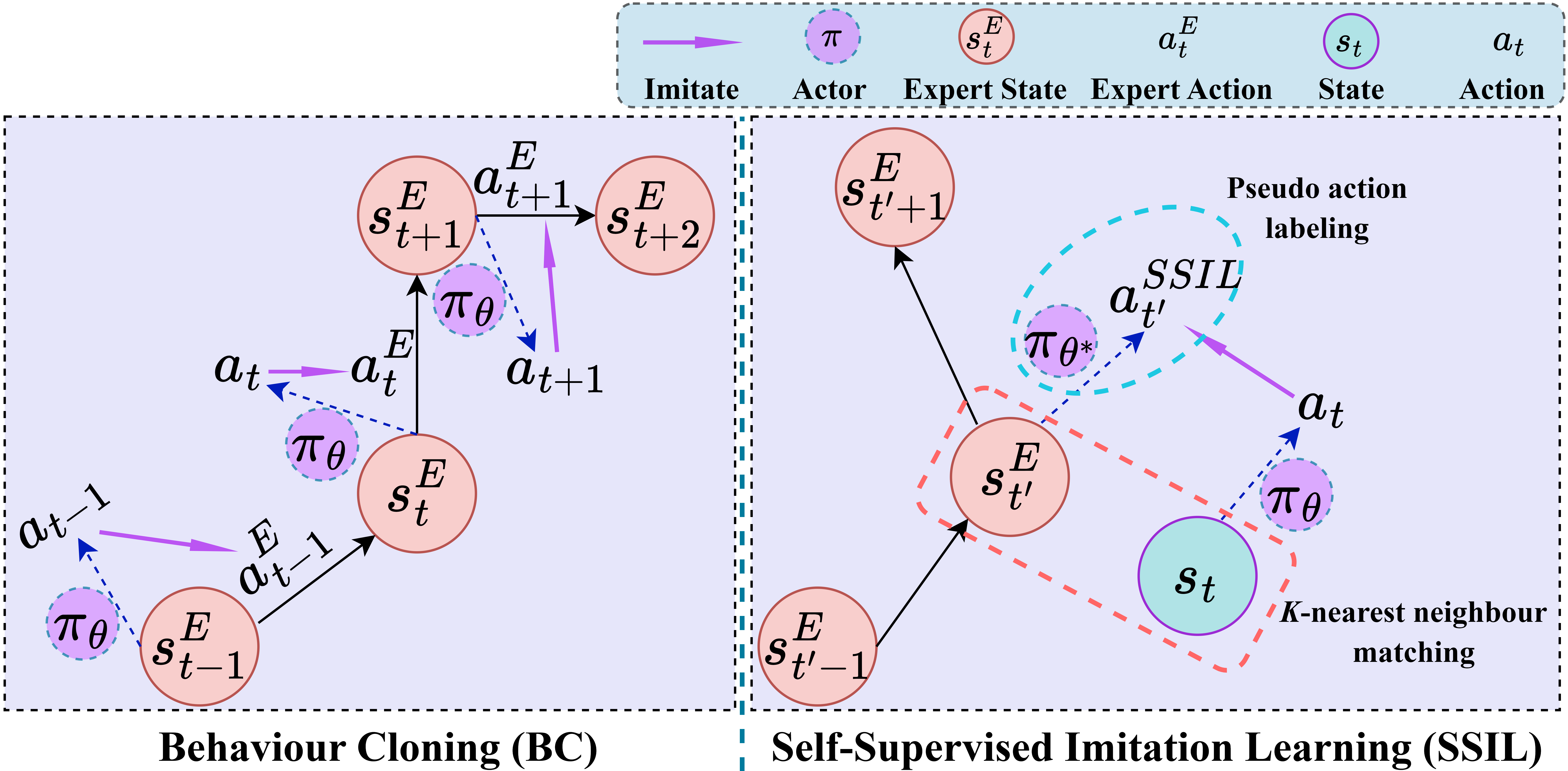}}
    \end{minipage}
    \caption{Comparison between behaviour cloning (BC) and self-supervised imitation learning (SSIL). BC regularizes agent training by minimizing the distance between the policy action and the demonstrated action, where action labels are necessary. The proposed SSIL retrieves from demonstrated states the nearest neighbours of the query state $s_t$ and produces pseudo action labels for exploration guidance, overcoming the need for action annotations.}
    \label{fig:SIL_demo}
\end{figure}

As our expert demonstrations only contain state information, we need a mechanism to integrate such knowledge into the training process mapping states to actions.  
Hence, we introduce the self-supervised imitation learning method, termed SSIL and demonstrated in Fig.~\ref{fig:ACSIL_demo} and~\ref{fig:SIL_demo}. SSIL improves the exploration efficiency and leads to better performance.


\subsubsection{$K$-Nearest Neighbour Matching}

Given a query state from the replay buffer, i.e. $s_t\in D_\pi$, we retrieve from the expert buffer $D_E$ its nearest neighbours via the Euclidean distance metric and use them to elicit actions to guide the policy learning process. We utilize the learned actor network to produce a pseudo action label for exploration guidance, which we expect can compete against methods that rely on action labels from demonstrations.

\subsubsection{Pseudo Action Labeling}

To guide the training of RL agents, potentially in the updates of both the actor and critic, the pseudo action label for a given query state $s_t$, dubbed $a^{SSIL}_{t'}$, is calculated as follows:
\begin{equation}
    a^{SSIL}_{t'}(s_t) = \mathbb{E}_{s_{t'}\in NN(s_t,D_E)}[\pi_{\theta^\ast}(s_{t'})],
    \label{eq:SIL_action}
\end{equation}
where $NN(s_t,D_E)$ denotes the set of the $K$-nearest neighbours of $s_t$ and $\pi_{\theta^\ast}$ refers to the target actor network. \footnote{We empirically found that using the current actor for action labeling can incur training instability and result in performance degradation.} The SSIL method produces guidance with state-only expert demonstrations in a self-supervised manner which jointly depends on the similarities to demonstrated states and the bootstrapping of policy agent. The pseudo actions are then leveraged to regularize the agent behaviour, as introduced in the next section.

\subsection{Actor-Critic SSIL Training Framework (AC-SSIL)}
\label{sect:AC_SIL}

We use the pseudo action labels produced by SSIL to regularize the learning process of RL models, aiming to incorporate knowledge from expert data into RL exploration.
The actor-critic SSIL training method, dubbed AC-SSIL and illustrated in Fig.~\ref{fig:ACSIL_demo}, is introduced here. Firstly, it adopts the RL update to retrieve optimal policies by learning from interactions with the environment. Secondly, it utilizes the SSIL method to leverage expert demonstrations into the RL paradigm to enhance model learning. The training steps are implemented based on a deep deterministic policy gradient framework~\cite{ddpg} with the replay buffer $D_\pi$ and the expert buffer $D_E$ for network update. The objective functions for the actor and critic networks are present in the following.

\subsubsection{Behaviour Regularized Actor Objective}

The training of the actor network is implemented by maximizing the action value estimated by the critic network and minimizing the distance between the demonstrated action elicited via the SSIL method and the policy action, yielding:
\begin{equation}
    \pi_\theta = \arg\max\limits_{\pi} \mathbb{E}_{s_t}[Q_\phi^\pi(s_t, \pi(s_t)) - \alpha d(\pi(s_t), a^{SSIL}_{t'}(s_t))], 
    \label{eq:actor_SIL}
\end{equation}
where $a^{SSIL}_{t'}$ is the pseudo action label of the query state $s_t$ and computed via~\eqref{eq:SIL_action}, $d(.,.)$ refers to the Euclidean distance that measures the similarity between actions, and $\alpha$ is the weight which balances the strengths of the RL and SSIL terms. The actor objective leverages both the action value estimation and imitation learning from state-only demonstrations in aid of exploring optimal solutions.

\subsubsection{Behaviour Regularized Critic Objective}

To overcome the overestimation problem of action value approximation which can be caused by error accumulation when the policy actions are out-of-distribution for the critic~\cite{DRL_Double_Q,TD3,offpolicy_batch_constrained,offline_RL_Fisher_critic}, the SSIL term is added to the target $Q$-value to advise the critic of guidance information from expert demonstrations. The objective function for the critic is therefore given by:
\begin{align}
    \hat{Q}(s_t, a_t) \coloneqq r_t + \gamma \mathbb{E}_{a'\sim \pi_{\theta^\ast}}[Q_{\phi^\ast}^\pi&(s_{t+1}, a') \notag\\
    &- \alpha d(a', a^{SSIL}_{t'+1}(s_{t+1}))] \notag\\
    Q_\phi^\pi = \arg\min\limits_{Q^\pi} \mathbb{E}_{(s_t,a_t)}[(\hat{Q}(s_t, a_t&) - Q^\pi(s_t, a_t))^2],
    \label{eq:critic_SIL}
\end{align}
which mitigates the over-estimation problem of ordinary critic that is inimical to agent exploration, by reducing the values of undesired actions through the distance regularization and thus propagating guidance from demonstrated states.

\begin{figure}[thpb]
    \begin{minipage}[b]{1.0\linewidth}
        \centering
        \centerline{\includegraphics[width=1.0\linewidth]{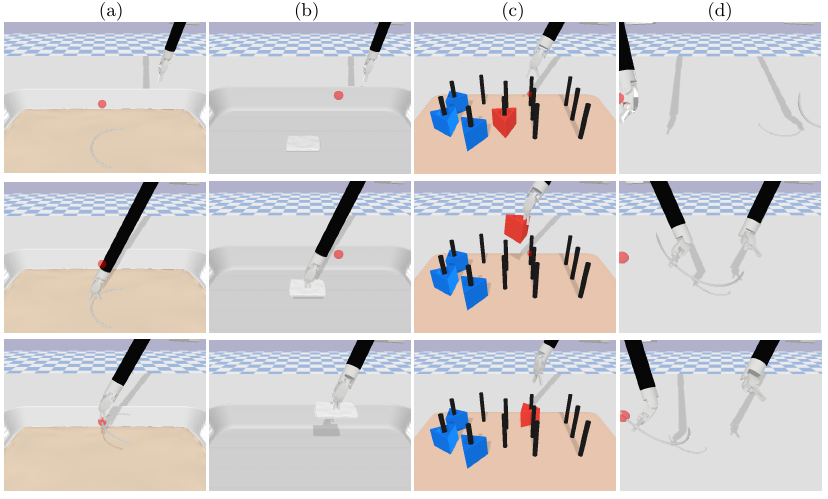}}
    \end{minipage}
    \caption{Surgical manipulation tasks automatically performed by RL agents: (a) NeedlePick, (b) GauzeRetrieve, (c) PegTransfer, and (d) NeedleRegrasp.}
    \label{fig:demo_surrol}
\end{figure}

\section{Experiments}

\subsection{Experiment Configurations}

We conduct experiments in an open-source simulation platform, dubbed SurRol~\cite{SurRol}, which is designed for surgical robot learning and provides manipulation tasks with varying degrees of complexity to facilitate relevant research. The transferability of the simulated environment was validated in~\cite{SurRol,dex}, showing that the agents trained in SurRol can be transferred to accomplish tasks in a real-world da Vinci Research Kit (dVRK) platform. We evaluate the model performance on four manipulation tasks as illustrated in Fig.~\ref{fig:demo_surrol}. These tasks have been selected because they encompass a broad variety of manipulation skills and exhibit varying levels of complexity, which is beneficial to demonstrate the capacity and dexterity of agent models to undertake surgical tasks. The tasks are:
1) \textit{NeedlePick}: approach and pick a needle using a robot arm, 2) \textit{GauzeRetrieve}: retrieve and pick a suture gauze and sequentially place it at the target position, 3) \textit{PegTransfer}: pick an object from one peg and move it to the target peg, and 4) \textit{NeedleRegrasp}: hand over the held needle from one robot arm to the other. All tasks are goal-conditioned with sparse reward functions indicating success. We adopt object state including 3D Cartesian positions and 6D pose and robot proprioceptive state including jaw status and end-effector position as state representation, and use Cartesian-space control as action space. 

In our experiments, the parameterized actor and critic networks are composed of four fully-connected layers of hidden dimension 256 with ReLU activations in between. The training procedure is actuated using an ADAM optimizer~\cite{adam} with $\beta_1$=0.9, $\beta_2$=0.999, and a learning rate of $10^{-3}$. We empirically set $K=5$ and $\alpha=5$ and sample 100 successful episodes as demonstrations via the expert policy provided by the environment SurRol, and assess the manipulation performance of different methods on each task after 150K environment steps as we found that increasing the training time did not remarkably enhance the model capacity.

\begin{table*}[thpb]
    \vspace{8pt}
    \caption{Comparison Results Where the Mean and Standard Deviation are Reported. Those Methods Leverage Demonstrations that either are Composed of Pure States (State-Only) or must Contain Action Labels (Action-Based).}
    \label{table:comparison}
    \centering
    \begin{minipage}[b]{1.0\linewidth}
        \centering
        \begin{tabularx}{0.79\linewidth}{|l||*{6}{c|}} 
            \toprule
            \diagbox{task}{method} & AMP~\cite{AMP} & DDPGBC~\cite{DDPGBC} & AWAC~\cite{awac}  & CoL~\cite{CoL} & DEX~\cite{dex} & AC-SSIL(ours) \\
            \midrule
            demo guidance & state-only & action-based & action-based  & action-based & action-based & state-only \\ 
            \midrule
            NeedlePick   & 0.77($\pm$0.10) & 0.96($\pm$0.02)  & 0.96($\pm$0.02)  &  \textbf{0.99}($\pm$0.02)  & \textbf{0.99}($\pm$0.02) & \textbf{0.99}($\pm$0.02)  \\
            GauzeRetrieve   & 0.50($\pm$0.07) &  0.68($\pm$0.08)  &  0.88($\pm$0.05)  & \textbf{0.89}($\pm$0.07)   & 0.85($\pm$0.06) & \textbf{0.89}($\pm$0.04)  \\
            PegTransfer   & 0.05($\pm$0.04) &  0.24($\pm$0.10)  &  0.48($\pm$0.19)  & 0.85($\pm$0.10)   & \textbf{0.95}($\pm$0.05) & 0.94($\pm$0.04)  \\
            NeedleRegrasp  & 0.04($\pm$0.03) &  0.16($\pm$0.06)  &  0.18($\pm$0.08)  & 0.22($\pm$0.09)   & 0.77($\pm$0.11) & \textbf{0.83}($\pm$0.08)   \\
            \bottomrule
        \end{tabularx}
    \end{minipage}
\end{table*}

\begin{figure}[thpb]
    \begin{minipage}[b]{1.0\linewidth}
        \centering
        \centerline{\includegraphics[width=1.0\linewidth]{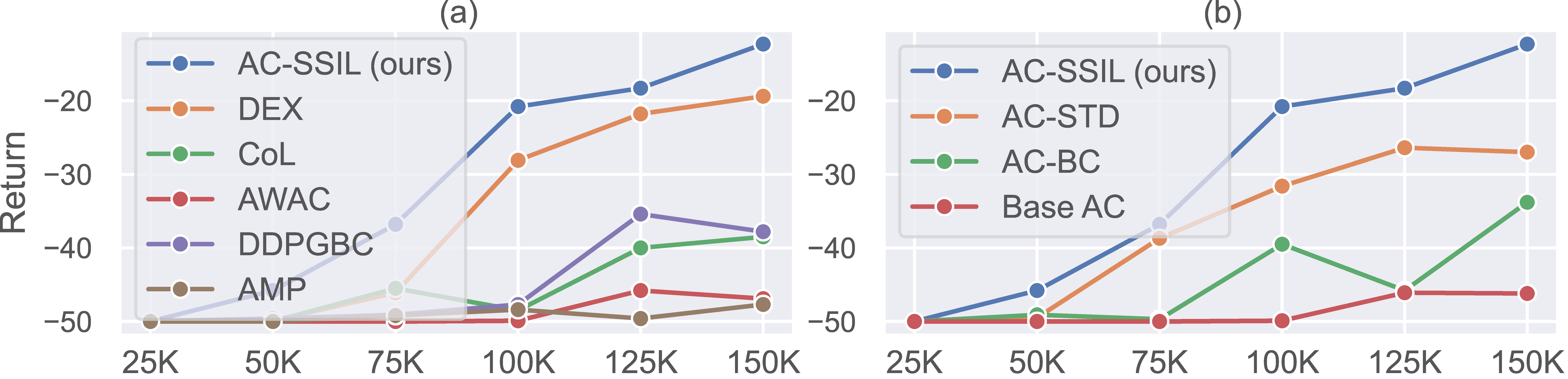}}
    \end{minipage}
    \caption{Evolution of return over training on NeedleRegrasp task. Our method AC-SSIL is compared against (a) methods in comparison and (b) methods in analysis on SSIL.}
    \label{fig:evolution}
\end{figure}

\subsection{Comparison Results}

We want to evaluate the performance of the AC-SSIL method which only relies on states to guide agent learning and compare it with action-based imitation learning methods.

\textbf{Setup} We compare with a method, dubbed AMP~\cite{AMP}, which was proposed to leverage state-only demonstrations and present comparisons with methods for integrating demonstrated actions into agent training, including DDPGBC~\cite{DDPGBC}, AWAC~\cite{awac}, CoL~\cite{CoL}, and DEX~\cite{dex}. The methods we compare with are briefly summarized below:
\begin{itemize}
    \item AMP~\cite{AMP} that extends GAIL~\cite{GAIL} to state-only demonstrations and adversarially imitates the expert behaviour using a discriminator to refine the reward;
    \item DDPGBC~\cite{DDPGBC} that regularizes the actor objective with a $Q$-filtered BC loss;
    \item AWAC~\cite{awac} that pre-trains an RL agent with demonstrations offline and fine-tunes it online with a $Q$-value based conservative constraint on policy learning;
    \item CoL~\cite{CoL} that adopts as initialization the agent pre-trained with BC offline and incorporates the BC regression into RL to maintain model performance and prevent the brittle degradation as training continues;
    \item DEX~\cite{dex} that propagates the demonstration guidance to the actor and critic updates during training by utilizing both states and actions from demonstrations.
\end{itemize}

Evaluations are over 10 runs with different random seeds, where each averages 20 trials with environment variations, e.g. the initial and target positions of the needle are varying. 

\textbf{Results} The comparison results are summarized in Table~\ref{table:comparison}, where it can be found that our method achieves competitive or better performance on different surgical tasks and surpasses all comparison methods on the challenging NeedleRegrasp task. It is also shown that the method AMP based on adversarial training leads to inferior performance, for it can suffer from the issues of training instability, inaccurate discrimination of states, and limited generalization~\cite{GAIL,interpretable_GAIL, perspective_IL}. The evolution of return over training is present in Fig.~\ref{fig:evolution} (a), which showcases the benefits of our method in enhancing model learning using demonstrations consisting of pure states to facilitate RL exploration.

\begin{figure*}[ht]
    \begin{minipage}[b]{1.0\linewidth}
        \vspace{6pt}
        \centering
        \centerline{\includegraphics[width=1.0\linewidth]{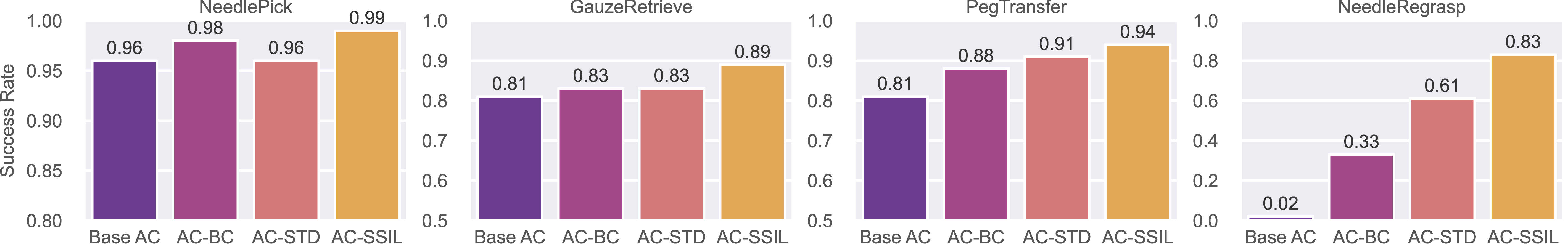}}
    \end{minipage}
    \caption{Analysis results using the actor-critic baseline~\cite{ddpg} (Base AC), the actor-critic methods with behaviour cloning~\cite{DDPGBC} (AC-BC) and with imitation learning via the state transition distance~\cite{OT_offlineRL,seabo} (AC-STD), and the proposed AC-SSIL.}
    \label{fig:SIL_analysis}
\end{figure*}

\subsection{Analysis on Self-Supervised Imitation Learning}
\label{sect:analysis_SIL}

We now want to  analyze and verify the efficacy of the proposed self-supervised imitation learning approach. 

\textbf{Setup} We compare with the baseline model, termed Base AC, which implements the actor-critic RL framework without using SSIL and expert demonstrations for training guidance~\cite{ddpg}, and the model, termed AC-BC, which leverages expert data by combining behaviour cloning~\cite{DDPGBC} with RL and inevitably requires action labels. Additionally, we also compare with a state-based IL method, termed AC-STD, which regularizes the action value estimation via the distance between the state transition experienced by the agent and the nearest neighbour in demonstrations~\cite{OT_offlineRL,seabo}.

\textbf{Results} The comparison results are present in Fig.~\ref{fig:SIL_analysis}, where it is evident that the proposed SSIL significantly improves the model performance on all tasks compared to the baseline model, e.g. the success rate on PegTransfer task rises from 0.81 to 0.94, and incurs more prominent improvements over behaviour cloning which can exhibit poor generalization capability on complex tasks, e.g. the success rate is enhanced from 0.33 to 0.83 on NeedleRegrasp task. Our method can consistently outperform the AC-STD that fails to directly steer the actor behaviour, which demonstrates the capacity of SSIL to effectively guide the agent training only using state observations. The evolution of return in Fig.~\ref{fig:evolution} (b) confirms the advantage of SSIL over other candidates. Those findings validate the advances of our method for leveraging state-only demonstrations to improve RL exploration.

\begin{figure}[thpb]
    \begin{minipage}[b]{1.0\linewidth}
        \centering
        \centerline{\includegraphics[width=1.0\linewidth]{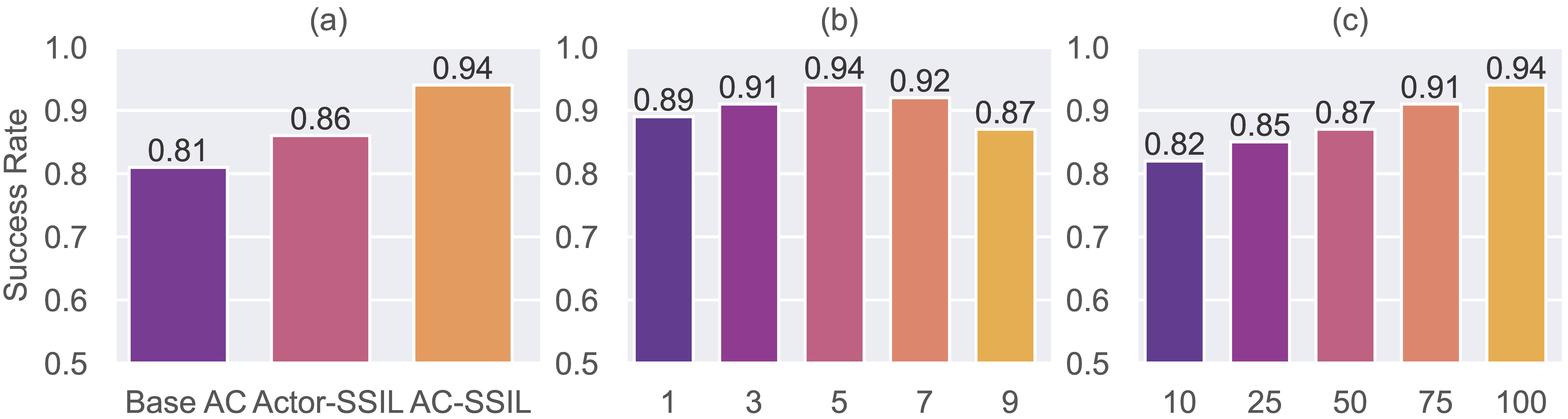}}
    \end{minipage}
    \caption{Ablation and sensitivity studies on PegTransfer task: (a) ablation on AC-SSIL training method, (b) sensitivity to the number $K$ of nearest neighbouts, and (c) sensitivity to expert demonstration amount.}
    \label{fig:SIL_ablate}
\end{figure}

\subsection{Ablation on AC-SSIL Training Method}
\label{sect:ablate}

To verify the effectiveness of the proposed AC-SSIL training framework introduced in Sect.~\ref{sect:AC_SIL}, we compare the Base AC and AC-SSIL models with the variant, termed Actor-SSIL, which only exploits the SSIL to regularize the actor objective. It reveals from the comparison results in Fig.~\ref{fig:SIL_ablate} (a) that the incorporation of SSIL in Actor-SSIL delivers a considerable improvement over Base AC, which verifies the efficacy of SSIL in enhancing agent learning. The performance gap between Actor-SSIL and AC-SSIL indicates that the impediments of demonstration guidance of ordinary critic has an unfavourable impact on model performance, which can be alleviated by regularizing the critic objective with the SSIL term. Those observations validate the effectiveness of the AC-SSIL training method.

\subsection{Sensitivity Analysis}
\label{sect:sensitivity}

We investigate the influences of the number $K$ of nearest neighbours and the amount of expert demonstrations. The results in Fig.~\ref{fig:SIL_ablate} (b) and (c) show that an intermediate value of $K$ around 5 works well and too large or small values can lead to a slight performance drop. Our method exhibits stable performance over an appropriate range of demonstration amount and more expert demonstrations can consistently enhance model performance. Those findings verify its insensitivity and robustness to those parameters.

\section{Conclusion}

An RL exploration framework is introduced in this work for the automation of surgical tasks, which can potentially assist surgeons in making informed decisions during the surgical operations. To enhance agent exploration with demonstrations that consist of pure state observations, thus making it applicable to expert data resources only with state information, a novel self-supervised imitation learning method is devised and utilized to guide the training processes. The improved performance over the baseline RL model showcases the effectiveness of our method, e.g. the success rates on PegTransfer and NeedleRegrasp tasks have been considerably enhanced from 0.81/0.02 to 0.94/0.83. The comparison results indicate that the method can compete with approaches that require action labels for behaviour imitation. Appealing research directions include extending our method to offline RL configurations and scaling it for more complex long-horizon tasks such as wound suturing, which can validate its usefulness and generalization to various artificial intelligence-assisted surgical operation scenarios including surgical training, planning and rehearsal.


\bibliographystyle{IEEEtran}
\bibliography{manuscript}

\begin{thebibliography}{10}
\providecommand{\url}[1]{#1}
\csname url@samestyle\endcsname
\providecommand{\newblock}{\relax}
\providecommand{\bibinfo}[2]{#2}
\providecommand{\BIBentrySTDinterwordspacing}{\spaceskip=0pt\relax}
\providecommand{\BIBentryALTinterwordstretchfactor}{4}
\providecommand{\BIBentryALTinterwordspacing}{\spaceskip=\fontdimen2\font plus
\BIBentryALTinterwordstretchfactor\fontdimen3\font minus
  \fontdimen4\font\relax}
\providecommand{\BIBforeignlanguage}[2]{{%
\expandafter\ifx\csname l@#1\endcsname\relax
\typeout{** WARNING: IEEEtran.bst: No hyphenation pattern has been}%
\typeout{** loaded for the language `#1'. Using the pattern for}%
\typeout{** the default language instead.}%
\else
\language=\csname l@#1\endcsname
\fi
#2}}
\providecommand{\BIBdecl}{\relax}
\BIBdecl

\bibitem{RL_book}
R.~S. Sutton and A.~G. Barto, \emph{{Reinforcement Learning: An Introduction}},
  2nd~ed.\hskip 1em plus 0.5em minus 0.4em\relax MIT Press, 2018.

\bibitem{deep_RL}
A.~Plaat, \emph{{Deep Reinforcement Learning}}.\hskip 1em plus 0.5em minus
  0.4em\relax Springer, 2022.

\bibitem{RL_surgery}
S.~Datta, Y.~Li, M.~Ruppert, and et~al, ``{Reinforcement Learning in
  Surgery},'' \emph{Surgery}, vol. 170,1, 2021.

\bibitem{RL_surgical_decision_making}
T.~J. Loftus, A.~C. Filiberto, Y.~Li, and et~al, ``{Decision Analysis and
  Reinforcement Learning in Surgical Decision-Making},'' \emph{Surgery}, vol.
  168, no.~2, pp. 273--278, 2020.

\bibitem{survey_RL_medical}
T.~Nguyen, I.~Nahum-Shani, E.~Wetter, and D.~Almirall, ``{Reinforcement
  Learning in Medical Decision Making: Applications, Challenges, and
  Opportunities},'' \emph{Journal of Biomedical Informatics}, vol. 120, p.
  103849, 2021.

\bibitem{alexnet}
A.~Krizhevsky, I.~Sutskever, and G.~E. Hinton, ``{ImageNet classification with
  deep convolutional neural networks},'' \emph{Advances in Neural Information
  Processing Systems}, vol.~25, 2012.

\bibitem{vgg}
K.~Simonyan and A.~Zisserman, ``{Very deep convolutional networks for
  large-scale image recognition},'' \emph{ICLR}, 2015.

\bibitem{vallina_gan}
I.~J. Goodfellow, J.~Pouget, M.~Mirza, and et~al, ``{Generative adversarial
  networks},'' \emph{NeurIPS}, vol.~27, pp. 2672--2680, 2014.

\bibitem{vit}
A.~Dosovitskiy, L.~Beyer, A.~Kolesnikov, and et~al, ``{An image is worth 16x16
  words: transformers for image recognition at scale},'' \emph{International
  Conference on Learning Representations}, 2021.

\bibitem{swin_transformer}
Z.~Liu, Y.~Lin, Y.~Cao, and et~al, ``{Swin transformer: hierarchical vision
  transformer using shifted windows},'' \emph{ICCV}, 2021.

\bibitem{DRL_robot_train}
X.~Tan, C.-B. Chng, Y.~Su, and et~al, ``{Robot-Assisted Training in Laparoscopy
  Using Deep Reinforcement Learning},'' \emph{IEEE Robotics and Automation
  Letters}, vol.~4, no.~2, pp. 485--492, 2019.

\bibitem{RL_robot_sugery}
A.~T. Bourdillon, A.~Garg, H.~Wang, and et~al, ``{Integration of Reinforcement
  Learning in a Virtual Robotic Surgical Simulation},'' \emph{Surgical
  innovation}, pp. 94--102, 2023.

\bibitem{sim2real_autonomous}
Y.~Ou and M.~Tavakoli, ``{Sim-to-Real Surgical Robot Learning and Autonomous
  Planning for Internal Tissue Points Manipulation Using Reinforcement
  Learning},'' \emph{IEEE Robotics and Automation Letters}, vol.~8, no.~5, pp.
  2502--2509, 2023.

\bibitem{sim2real_CycleGAN}
P.~M. Scheikl, E.~Tagliabue, B.~Gyenes, and et~al, ``{Sim-to-Real Transfer for
  Visual Reinforcement Learning of Deformable Object Manipulation for
  Robot-Assisted Surgery},'' \emph{IEEE Robotics and Automation Letters},
  vol.~8, no.~2, pp. 560--567, 2023.

\bibitem{loop_healthcare}
Y.~Dai, G.~Wang, K.~Muhammad, and S.~Liu, ``{A Closed-Loop Healthcare
  Processing Approach Based on Deep Reinforcement Learning},'' \emph{Multimedia
  Tools Appl.}, vol.~81, no.~3, p. 3107–3129, jan 2022.

\bibitem{general_RL_Med}
S.~Schmidgall, J.~W. Kim, A.~Kuntz, A.~E. Ghazi, and A.~Krieger,
  ``{General-purpose foundation models for increased autonomy in robot-assisted
  surgery},'' 2024.

\bibitem{GAIL_RL_robot}
Y.~Ou and M.~Tavakoli, ``Towards safe and efficient reinforcement learning for
  surgical robots using real-time human supervision and demonstration,''
  \emph{ISMR}, pp. 1--7, 2023.

\bibitem{DDPGBC}
A.~Nair, B.~McGrew, M.~Andrychowicz, W.~Zaremba, and P.~Abbeel, ``{Overcoming
  Exploration in Reinforcement Learning with Demonstrations},'' \emph{ICRA},
  2018.

\bibitem{CoL}
M.~Vecerik, C.~Hesse, T.~Wang, and et~al, ``{Integrating Behavior Cloning and
  Reinforcement Learning for Improved Performance in Dense and Sparse Reward
  Environments},'' \emph{Advances in Neural Information Processing Systems},
  pp. 5308--5317, 2017.

\bibitem{minimalist_offline}
S.~Fujimoto and S.~S. Gu, ``{A Minimalist Approach to Offline Reinforcement
  Learning},'' \emph{NeurIPS}, vol.~34, pp. 15\,339--15\,352, 2021.

\bibitem{revisit_minimalist_offline}
D.~Tarasov, V.~Kurenkov, A.~Nikulin, and S.~Kolesnikov, ``{Revisiting the
  Minimalist Approach to Offline Reinforcement Learning},'' \emph{Advances in
  Neural Information Processing Systems 36 (NeurIPS)}, 2023.

\bibitem{survey_IL}
A.~Hussein, M.~M. Gaber, E.~Elyan, and C.~Jayne, ``{Imitation Learning: A
  Survey of Learning Methods},'' \emph{ACM Computing Surveys (CSUR)}, vol.~50,
  no.~2, pp. 1--35, 2017.

\bibitem{survey_demo_robot}
B.~D. Argall, S.~Chernova, M.~Veloso, and B.~Browning, ``{A Survey of Robot
  Learning from Demonstration},'' \emph{Robotics and Autonomous Systems},
  vol.~57, no.~5, pp. 469--483, 2009.

\bibitem{IL_Obs_survel}
F.~Torabi, G.~Warnell, and P.~Stone, ``{Recent Advances in Imitation Learning
  from Observation},'' \emph{IJCAI}, pp. 6325--6331, 7 2019.

\bibitem{RCE}
B.~Eysenbach, S.~Levine, and R.~Salakhutdinov, ``{Replacing Rewards with
  Examples: Example-Based Policy Search via Recursive Classification},''
  \emph{Advances in Neural Information Processing Systems}, 2021.

\bibitem{LobsDICE}
G.-H. Kim, J.~Lee, Y.~Jang, H.~Yang, and K.-E. Kim, ``{LobsDICE: Offline
  Learning from Observation via Stationary Distribution Correction
  Estimation},'' \emph{NeurIPS}, 2022.

\bibitem{GAIL}
J.~Ho and S.~Ermon, ``{Generative Adversarial Imitation Learning},''
  \emph{Advances in Neural Information Processing Systems}, vol.~29, 2016.

\bibitem{GAIfO}
F.~Torabi, G.~Warnell, and P.~Stone, ``Generative adversarial imitation from
  observation,'' \emph{ICML Workshop}, June 2019.

\bibitem{AMP}
X.~B. Peng, Z.~Ma, P.~Abbeel, S.~Levine, and A.~Kanazawa, ``{AMP: Adversarial
  Mtion Priors for Stylized Physics-Based Character Control},'' \emph{ACM
  Trans. Graph.}, vol.~40, no.~4, jul 2021.

\bibitem{seabo}
Z.~Yuan, D.~Wu, X.~Lin, M.~Lin, Z.~Zhang, S.~Ma, H.~Zhang, and B.~Zhou,
  ``{SEABO: A Simple Search-Based Method for Offline Imitation Learning},''
  \emph{NeurIPS}, pp. 18\,032--18\,042, 2020.

\bibitem{interpretable_GAIL}
W.~Liu, D.~Li, E.~Aasi, R.~Tron, and C.~Belta, ``{Interpretable Generative
  Adversarial Imitation Learning},'' \emph{CoRR}, vol. abs/2402.10310, 2024.

\bibitem{survey_IRL}
S.~Arora, P.~Doshi, and H.~Younger, ``{A Survey of Inverse Reinforcement
  Learning: Challenges, Methods and Progress},'' \emph{Artificial Intelligence
  Review}, vol.~55, p. 4307–4346, 2022.

\bibitem{SIL}
J.~Oh, Y.~Guo, S.~Singh, and H.~Lee, ``{Self-Imitation Learning},''
  \emph{ICML}, 2018.

\bibitem{human_loop_surgical_sim}
Y.~Long, W.~Wei, T.~Huang, Y.~Wang, and Q.~Dou, ``{Human-in-the-loop Embodied
  Intelligence with Interactive Simulation Environment for Surgical Robot
  Learning},'' \emph{RAL}, 2023.

\bibitem{RL_robot_inverse}
H.~Su, Y.~Hu, Z.~Li, and et~al, ``{Reinforcement Learning Based Manipulation
  Skill Transferring for Robot-Assisted Minimally Invasive Surgery},''
  \emph{ICRA}, pp. 2203--2208, 2020.

\bibitem{multi_agent_robot_surgery}
P.~M. Scheikl, B.~Gyenes, T.~Davitashvili, and et~al, ``{Cooperative Assistance
  in Robotic Surgery through Multi-Agent Reinforcement Learning},''
  \emph{IROS}, pp. 1859--1864, 2021.

\bibitem{IRL_steer_needle}
A.~Segato, M.~D. Marzo, S.~Zucchelli, S.~Galvan, R.~Secoli, and E.~De~Momi,
  ``{Inverse Reinforcement Learning Intra-Operative Path Planning for Steerable
  Needle},'' \emph{IEEE Transactions on Biomedical Engineering}, vol.~69,
  no.~6, pp. 1995--2005, 2022.

\bibitem{organ_context_RL}
C.~D.Ettorre, S.~Zirino, N.~N.Dei, A.~Stilli, E.~Momi, and D.~Stoyanov,
  ``{Learning Intraoperative Organ Manipulation with Context-Based
  Reinforcement Learning},'' \emph{International Journal of Computer Assisted
  Radiology and Surgery}, 2022.

\bibitem{RL_needle_pick}
R.~Bendikas, V.~Modugno, D.~Kanoulas, F.~Vasconcelos1, and D.~Stoyanov1,
  ``{Learning Needle Pick-And-Place without Expert Demonstrations},''
  \emph{IEEE Robotics and Automation Letters}, 2023.

\bibitem{ViSkill}
T.~Huang, K.~Chen, W.~Wei, J.~Li, Y.~Long, and Q.~Dou, ``{Value-Informed Skill
  Chaining for Policy Learning of Long-Horizon Tasks with Surgical Robot},''
  \emph{IROS}, 2023.

\bibitem{SQIL}
S.~Reddy, A.~D. Dragan, and S.~Levine, ``{SQIL: Imitation Learning via
  Reinforcement Learning with Sparse Rewards},'' \emph{ICLR}, 2020.

\bibitem{driving_online_IL}
Y.~Pan, C.-A. Cheng, K.~Saigol, and et~al, ``{Imitation Learning for Agile
  Autonomous Driving},'' \emph{The International Journal of Robotics Research},
  vol.~39, no. 2-3, pp. 286--302, 2020.

\bibitem{awac}
S.~Fujimoto, D.~Meger, and D.~Precup, ``{AWAC: Accelerating Online
  Reinforcement Learning with Offline Datasets},'' \emph{ICML}, 2020.

\bibitem{dex}
T.~Huang, K.~Chen, B.~Li, and et~al, ``{Demonstration-Guided Reinforcement
  Learning with Efficient Exploration for Task Automation of Surgical Robot},''
  \emph{IEEE International Conference on Robotics and Automation (ICRA)}, 2023.

\bibitem{hybrid_IL}
E.~Jung and I.~Kim, ``{Hybrid Imitation Learning Framework for Robotic
  Manipulation Tasks},'' \emph{Sensors}, vol.~21, no.~10, 2021.

\bibitem{BC_obs}
F.~Torabi, G.~Warnell, and P.~Stone, ``{Behavioral Cloning from Observation},''
  \emph{IJCAI}, 2018.

\bibitem{IRL}
A.~Y. Ng and S.~J. Russell, ``{Algorithms for Inverse Reinforcement
  Learning},'' \emph{ICML}, pp. 663--670, 2000.

\bibitem{entropy_IRL}
B.~D. Ziebart, A.~L. Maas, J.~A. Bagnell, and A.~K. Dey, ``{Maximum Entropy
  Inverse Reinforcement Learning},'' \emph{AAAI Conference on Artificial
  Intelligence (AAAI-08)}, pp. 1433--1438, 2008.

\bibitem{OT_offlineRL}
K.~Wei, M.~J. Bansal, S.~M. Kakade, C.~Daskalakis, and A.~A. Rusu, ``{Optimal
  Transport for Offline Imitation Learning},'' \emph{Advances in Neural
  Information Processing Systems}, pp. 14\,694--14\,704, 2020.

\bibitem{HER}
M.~Andrychowicz, D.~Crow, A.~Ray, and et~al, ``{Hindsight Experience Replay},''
  \emph{NIPS}, pp. 5048--5058, 2017.

\bibitem{DQN}
V.~Mnih, K.~Kavukcuoglu, D.~Silver, and et~al, ``{Human-Level Control Through
  Deep Reinforcement Learning},'' \emph{Nature}, vol. 518, no. 7540, pp.
  529--533, 2015.

\bibitem{q_learning}
C.~J. Watkins and P.~Dayan, ``{Q-Learning},'' \emph{Machine Learning}, vol.~8,
  no. 3-4, pp. 279--292, 1992.

\bibitem{RL_survey_robot}
J.~Kober, J.~A. Bagnell, and J.~Peters, ``{Reinforcement Learning in Robotics:
  A Survey},'' \emph{The International Journal of Robotics Research}, vol.~32,
  no.~11, pp. 1238--1274, 2013.

\bibitem{TD3}
S.~Fujimoto, H.~van Hoof, and D.~Meger, ``{Addressing Function Approximation
  Error in Actor-Critic Methods},'' \emph{ICML}, vol.~80, pp. 1587--1596, 2018.

\bibitem{TATD3}
T.~Joshi, S.~Makker, H.~Kodamana, and H.~Kandath, ``{Twin actor Twin Delayed
  Deep Deterministic Policy Gradient (TATD3) Learning for Batch Process
  Control},'' \emph{Computers \& Chemical Engineering}, vol. 155, p. 107527,
  2021.

\bibitem{ddpg}
T.~P. Lillicrap, J.~J. Hunt, A.~Pritzel, N.~Heess, T.~Erez, Y.~Tassa,
  D.~Silver, and D.~Wierstra, ``{Continuous Control with Deep Reinforcement
  Learning},'' \emph{ICLR May 2-4}, 2016.

\bibitem{DRL_Double_Q}
H.~v. Hasselt, A.~Guez, and D.~Silver, ``{Deep reinforcement learning with
  double Q-Learning},'' \emph{AAAI}, p. 2094–2100, 2016.

\bibitem{offpolicy_batch_constrained}
S.~Fujimoto, D.~Meger, and D.~Precup, ``{Off-Policy Deep Reinforcement Learning
  without Exploration},'' \emph{ICML}, 2018.

\bibitem{offline_RL_Fisher_critic}
I.~Kostrikov, R.~Fergus, J.~Tompson, and O.~Nachum, ``{Offline Reinforcement
  Learning with Fisher Divergence Critic Regularization},'' \emph{Proceedings
  of the 38th International Conference on Machine Learning, 18-24 July 2021},
  vol. 139, pp. 5774--5783, 2021.

\bibitem{SurRol}
J.~Xu, B.~Li, B.~Lu, and et~al, ``{SurRoL: An Open-source Reinforcement
  Learning Centered and dVRK Compatible Platform for Surgical Robot
  Learning},'' \emph{IROS}, 2021.

\bibitem{adam}
D.~P. Kingma and J.~Ba, ``{Adam: a method for stochastic optimization},''
  \emph{International Conference on Learning Representations}, May 2015.

\bibitem{perspective_IL}
T.~Osa, J.~Pajarinen, and G.~Neumann, \emph{{An Algorithmic Perspective on
  Imitation Learning}}.\hskip 1em plus 0.5em minus 0.4em\relax Hanover, MA,
  USA: Now Publishers Inc., 2018.

\end{thebibliography}

\end{document}